\documentclass[11pt,a4paper]{article}
\usepackage[hyperref]{acl2019}
\usepackage[utf8]{inputenc}
\usepackage{graphicx}
\usepackage{subcaption}
\usepackage{xcolor}
\usepackage{footnote}

\makeatletter
\newcommand\footnoteref[1]{\protected@xdef\@thefnmark{\ref{#1}}\@footnotemark}
\makeatother

\title{Spoken Conversational Search for General Knowledge}

\author{Lina M.~Rojas-Barahona, Pascal Bellec, Benoit Besset, Martinho Dos-Santos, \\ \textbf{Johannes Heinecke,  Munshi Asadullah, Olivier Le-Blouch, Jean Y. Lancien, } \\ \textbf{Géraldine Damnati, Emmanuel Mory and Frédéric Herledan}\\
 Orange Labs, 2 Avenue de Pierre Marzin, Lannion, France \\
  \texttt{\{linamaria.rojasbarahona,pascal.bellec,benoit.besset,martinho.dossantos,}   \\\texttt{johannes.heinecke,munshi.asadullah,olivier.leblouch,jeanyves.lancien} 
  \\\texttt{geraldine.damnati,emmanuel.mory,frederic.herledan\}@orange.com} \\}
 
 \aclfinalcopy
\date{}

\begin{document}

\maketitle
\begin{abstract}
We present a spoken conversational question answering proof of concept that is able to answer questions about general knowledge from Wikidata\footnote{\label{wiki}\url{https://www.wikidata.org}}. The dialogue component does not only orchestrate various components but also solve coreferences and ellipsis.

\end{abstract}
\section{Introduction}
Conversational question answering is an open research problem. It studies the integration of \textit{question answering} (QA) systems in a \textit{dialogue system}(DS). Not long ago, each of these research subjects were studied separately; only very recently has studying the intersection between them gained increasing interest~\cite{reddy2018coqa,choi2018quac}.

We present a spoken conversational question answering system that is able to answer questions about general knowledge in French by calling two distinct QA systems. It solves coreference and ellipsis by modelling context. Furthermore, it is extensible, thus other components such as neural approaches for question-answering can be easily integrated. It is also possible to collect a dialogue corpus from its iterations.

In contrast to most conversational systems which support only speech, two input and output modalities are supported speech and text. Thus it is possible to let the user check the answers by either  asking relevant Wikipedia excerpts or by navigating through the retrieved name entities or by exploring the answer details of the QA components: the confidence score as well as the set of explored triplets. Therefore, the user has the final word to consider the answer as correct or incorrect and to provide a reward, which can be used in the future for training reinforcement learning algorithms. 

\section{Architectural Description}
The high-level architecture of the proposed system consists of a speech-processing front-end, an understanding component, a context manager, a generation component, and a synthesis component. The context manager provides contextualised mediation between the dialogue components and several question answering back-ends, which rely on data provided by Wikidata\footnoteref{wiki}. Interaction with a human user is achieved through a graphical user interface (GUI). Figure 1 depicts the components together with their interactions.


\begin{figure}[h!]
	\centerline{\includegraphics[width=\linewidth]{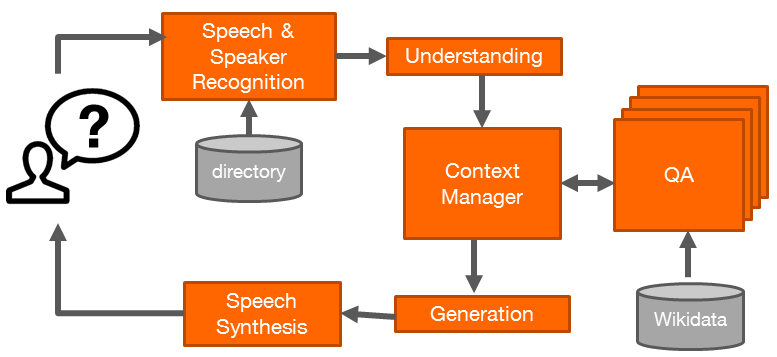}}
	\caption{\small\noindent High-level depiction of the proposed spoken conversation question answering system. Arrows indicate data flow and direction.}
	\label{f:achictecture}
\end{figure}

In the remainder of this section, we explain the components of our system.

\subsection{Speech and Speaker Recognition}

The user vocally asks her question which is recorded through a microphone driven by the GUI. The audio chunks are then processed in parallel by a speech recognition component and a speaker recognition component.

\paragraph{Speech Recognition} The Speech Recognition component enables the translation of speech into text. Cobalt Speech Recognition for French is a Kaldi-based speech-to-text decoder using a TDNN \cite{speech} acoustic model; trained on more than 2\,000 hours of clean and noisy speech, a 1.7-million-word lexicon, and a 5-gram language model trained on 3 billion words.
\paragraph{Speaker Recognition} The Speaker Recognition component answers the question ``Who is speaking?''. This component is based on deep neural network speaker embeddings called ``x-vectors'' \cite{xvectors}. Our team participated to the NIST SRE18 challenge \cite{nistsre18}, reaching the 11th position among 48 participants. 

Once identified, it is possible to access the information of the speaker by accessing a speaker database which includes attributes such as nationality. This is a key module for personalising the behaviour of the system, for instance, by supporting questions such as "Who is the president of the country where I was born?".

\subsection{The Dialogue System}
\label{ss:dial}
\begin{figure}[h!]
	\includegraphics[width=\linewidth]{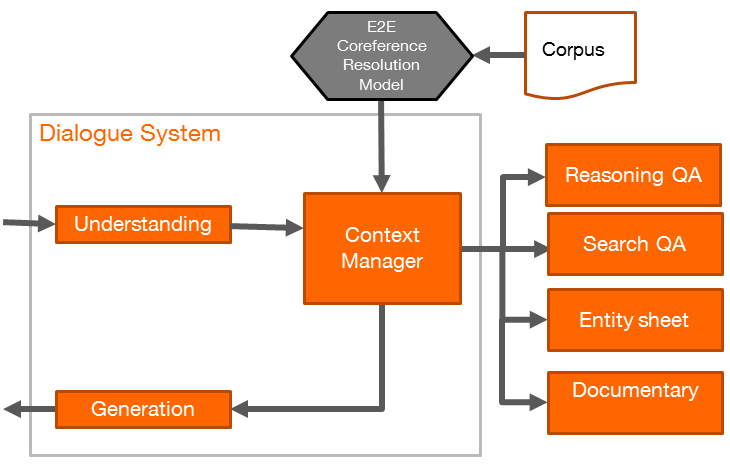}
	\caption{\small\noindent Internal structure of the proposed dialogue system, with emphasis placed on the interactions of the context manager.}
	\label{f:dialachictecture}
\end{figure}

The transcribed utterance and the speaker information are passed to the dialogue system. This system contains an \textbf{understanding} component,  a \textbf{context manager}, and a \textbf{generation} component (Figure~\ref{f:dialachictecture}). 

\paragraph{Understanding} 
The understanding component relies on a \textbf{linguistic module} to parser the user's inputs. The linguistic module supports part-of-the-speech (POS) tagging, lemmatisation, dependency syntax and semantics provided
by an adapted version of UDpipe~\cite{udpipe:2017}, extended with a French full-form lexicon. UDpipe was trained on the French GSD treebank version 2.3\footnote{\url{http://universaldependencies.org/}}. 
Since the syntax of questions in French differs from that of declaratives, we annotated manually about 500 questions to be merged into the UD treebank (which originally
did not contain questions). Tests show that the labelled attached score (LAS) is thereby increased by 10\% absolute, to 92\%.

\paragraph{Context Manager} The Context Manager component is able to solve coreferences by using an adaptation of the end-to-end model presented in ~\cite{lee2017end}, that we trained for French by using fasttext multilingual character embeddings~\cite{bojanowski2017enriching}.  The data used to train the coreference resolution model is a subset of the corpus CALOR~\cite{marzinotto2018semantic} (Table \ref{t:corefdata}), which has been manually annotated with coreferences. This corpus contains coreference chains of named entities, nouns and pronouns (such as ``the president'' -- ``JFK'' -- ``he'' -- ``his'').

\begin{table}[]
    \centering
    \begin{tabular}{l r r r}
    &\scriptsize{Train}&\scriptsize{Dev}&\scriptsize{Test}\\
    \hline
\scriptsize{words}                & \scriptsize{208\,245} & \scriptsize{45\,001} & \scriptsize{89\,330} \\[-1ex]
\scriptsize{sentences}~~~~        &  \scriptsize{10\,166} &  \scriptsize{2\,976} &  \scriptsize{4\,853} \\[-1ex]
\scriptsize{mentions}             &  \scriptsize{15\,013} &  \scriptsize{3\,008} &  \scriptsize{6\,232} \\[-1ex]
\scriptsize{~~~incl. prons.} &   \scriptsize{1\,465} &     \scriptsize{280} &     \scriptsize{538} \\[-1ex]
\scriptsize{chains}               &   \scriptsize{3\,793} &     \scriptsize{901} &  \scriptsize{1\,533} \\\hline
    \end{tabular}
    \caption{\small{Subset of the corpus CALOR used for training, developing and testing of the coreference resolution module. Note that the values given for the mentions include pronouns.}}
    \label{t:corefdata}
\end{table}


The dependency tree and semantic frames provided by the linguistic module are used to solve ellipsis by taking into account the syntactic and semantic structure of the previous question. 
Once the question has been resolved, it calls the QA systems and passes their results to the generation module.

\paragraph{Generation} 
The generation component either returns the short answer provided by QA systems or relies on an external generation module that uses dependency grammar templates to generate more elaborated answers. 

 \subsection{QA Systems}
Two complementary question answering components were integrated into the system: the Reasoning QA and Search QA. Each of these QA systems computes a confidence score for every answer by using icsiboost~\cite{icsiboost}, an Adaboost-based classifier trained on a corpus of around 21\,000 questions. The Context Manager takes into account these scores to pick the higher-confidence of the two answers. 

Besides the QA components, there are two other components that are able to provide complementary information about the Wikidata's entities under discussion: Documentary and Entity Sheet.
\paragraph{Reasoning QA}
The Reasoning QA system first parses the question by using a Prolog definite clause grammar (DCG), extended with word-embeddings to support variability in the vocabulary. Then it explores a graph containing logical patterns that are used to produce requests in SPARQL\footnote{\url{https://www.w3.org/TR/sparql11-query/}} that agree with the question.
\paragraph{Search QA}
The Search QA system uses an internal knowledge base, which finely indexes data using Elasticsearch. It is powered by Wikidata and enriched by Wikipedia, especially to calculate a Page-Rank ~\cite{pagelarrypagerank} on each entity.
This QA system first determines the potential named entities in the question (i.e. subjects, predicates, and types of subjects). Second, it constructs a correlation matrix by looking for the triplets in Wikidata that link these entities. This matrix is filtered according the coverage of the question and the relevance of each entity in order to find the best answer.


\paragraph{Documentary}
The documentary component is able to extract pertinent excerpts of Wikipedia. It uses an internal documentary base, which indexes Wikipedia's paragraphs by incorporating the Wikidata entity's IDs into elasticsearch indexes. Thus, it is possible to  find paragraphs (ranked by elasticsearch) illustrating the answer to the given question by taking into account the entities detected in the question and in the answer. 
\paragraph{Entity Sheet}
The entity sheet component summarises an entity in Wikidata returning the description, the picture and the type of the entity.

\subsection{Speech Synthesis}
Finally, the generated response is passed to the GUI, which in turn passes it to the Voxygen synthesis solution.

\section{Evaluation}
The evaluation of the individual components of the proposed system was performed outside the scope of this work. We evaluated out-of-context questions, as well as the coreference resolution module. 
\begin{figure}
\centerline{\includegraphics[scale=.37]{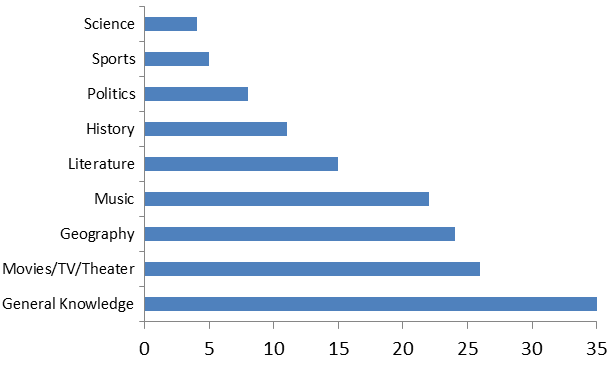}}
\caption{\small\noindent Distribution of question topics used to evaluate system performance on out-of-context questions.}
\label{f:testset}
\end{figure}


Performance on out-of-context questions was evaluated on Bench'It, a dataset containing 150 open ended questions about general knowledge in French (Figure~\ref{f:testset})\footnote{Publicly available in \url{https://github.com/lmrojasb/benchit.git}}. The system reached a macro precision, recall and F-1 of $64.14\%$, $64.33\%$ and $63.46\%$ respectively\footnote{Following the metrics of the task-4 of QALD-7 \url{https://project-hobbit.eu/challenges/qald2017/}}.

 We also evaluated the coreference resolution model on the test-set of CALOR (Table~\ref{t:corefdata}), obtaining an average precision, recall and F-1 of 65.59\%, 48.86\% and 55.77\% respectively. The same model reached a average F-1 of 68.8\% for English~\cite{lee2017end}. Comparable measurements are not available for French. F-1 scores for French are believed to be lower because of the lower amount of annotated data.






\section{Examples}
\begin{figure}[ht]
\noindent
\scriptsize{
\begin{tabular}{l}
    
    \colorbox{blue!20}{\scriptsize{U: Who is Michael Jackson ?}}\\[-0.5ex]
    \scriptsize{S: Michael Jackson is an American author,composer,}\\[-1ex]
    \scriptsize{singer and dancer}\\
    \colorbox{blue!20}{\scriptsize{U: What is his father's name?}}\\[-0.5ex]
    \scriptsize{S: Joseph Jackson}\\
    \colorbox{blue!20}{\scriptsize{U: and his mother's?}}\\[-0.5ex]
    \scriptsize{S: Katherine Jackson}\\
    \colorbox{blue!20}{\scriptsize{U: and his brothers' and sisters'?}}\\[-0.5ex]
    \scriptsize{S: Tito Jackson, Rebbie Jackson, Randy Jackson,}\\[-1ex]
     \scriptsize{Jackie Jackson, Marlon Jackson, La Toya Jackson,} \\[-1ex]
     \scriptsize{Jermaine Jackson,  Janet Jackson}
   
\end{tabular} }

\caption{\small\noindent English translation of French conversation involving in-context questions.}
\label{f:incontextex}
\end{figure}


 On the one hand, the system is able to answer complex out-of-context questions such as ``What are the capitals of the countries of the Iberian Peninsula?", by correctly answering the list of capitals: ``Andorra la Vella, Gibraltar, Lisbon, Madrid".
 
 On the other hand, consider the dialogue presented in Figure~\ref{f:incontextex}, in which the user asks several related questions about Michael Jackson. First she asks ``Who is Michael Jackson?'' and the system correctly answers ``Michael Jackson is an American author, composer, singer and dancer'', note that this is the generated long answer.
 
The subsequent questions are related to the names of his family members. In order to correctly answer these questions, the resolution of coreferences is neccesary to solve the possessive pronouns, which in French agree in gender and number with the noun they introduce. In this specific example, while in English ``his'' is used in all the cases, in French it changes to: \emph{son} père (father), \emph{sa} mère (mother), \emph{ses} frères (brothers).  This example also illustrates resolution of elliptical questions in the context, by solving the question ``and his mother's" as ``What is the name of his mother".



\section{Conclusion and Future Work}
\label{sec:conclusion}
We have presented a spoken conversational question answering system, in French. The DS orchestrates different QA systems and returns the response with the higher confidence score. The system contains modules specifically designed for dealing with common spoken conversation phenomena such as coreference and ellipsis. 

We will soon integrate a state-of-the art reading comprehension approach, support English language and
improve the coreference resolution module.
We are also interested in exploring policy learning, thus the system will be able to find the best criterion to chose the answer or to ask for clarification in the case of ambiguity and uncertainty. 
\bibliography{convsfgk}
\bibliographystyle{convsfgk}
\end{document}